\documentclass[runningheads]{llncs}
\usepackage{graphicx}
\usepackage{amsmath,amssymb} 
\usepackage{color}
\usepackage[width=122mm,left=12mm,paperwidth=146mm,height=193mm,top=12mm,paperheight=217mm]{geometry}

\usepackage{subfigure}
\usepackage{multirow}

\usepackage{algorithm}
\usepackage{algpseudocode}

\graphicspath{{./figures/}{./figures/draw/}}
\def\x{{\mathbf x}}

\def\w{{\mathbf w}}

\def\cc{{\mathbf c}}
\def\bb{{\mathbf b}}
\def\I{{\mathbf 1}}
\def\q{{\mathbf q}}
\def\B{{\mathbf B}}
\def\X{{\mathbf X}}
\def\Y{{\mathbf Y}}
\def\R{{\mathbb R}}
\def\H{{\mathbf H}}
\def\Z{{\mathbf Z}}
\def\W{{\mathbf W}}
\def\V{{\mathbf V}}
\def\Q{{\mathbf Q}}
\def\I{{\mathbf I}}
\def\1{{\mathbf 1}}
\def\S{{\mathbf S}}
\def\V{{\mathbf V}}
\newcommand\norm[1]{\left\lVert#1\right\rVert}

\begin{document}
\pagestyle{headings}
\mainmatter

\title{Learning to Hash with Binary Deep Neural Network} 

\titlerunning{\textit{Learning to Hash with Binary Deep Neural Network}}

\authorrunning{\textit{Thanh-Toan Do, Anh-Dzung Doan, Ngai-Man Cheung}}

\author{Thanh-Toan Do, Anh-Dzung Doan, Ngai-Man Cheung}


\institute{Singapore University of Technology and Design \\ {\tt\small \{thanhtoan\_do, dung\_doan, ngaiman\_cheung\}@sutd.edu.sg}}

\maketitle

\begin{abstract}

This work proposes deep network models and learning algorithms for unsupervised  and supervised binary hashing.
Our novel network design constrains one hidden layer to directly output the binary codes. 
This addresses a challenging issue in some previous works: optimizing non-smooth objective functions due to binarization.
Moreover, we incorporate independence and balance properties in the direct and strict forms in the learning. Furthermore, we  include  similarity preserving property in our objective function. 
Our resulting optimization with these binary, independence, and balance constraints is  difficult to solve. 
We propose to attack it with alternating optimization and careful relaxation. 
Experimental results on three benchmark datasets show that our proposed methods compare favorably with the state of the art.

\end{abstract}
\section{Introduction}
\vspace{-0.2cm}
We are interested in  learning binary hash codes for large scale visual search. Two main difficulties  with large scale visual search are efficient storage and fast searching. An attractive approach for handling these difficulties is binary hashing, where each original high dimensional vector $\x \in \R^D$ is mapped to a 
very compact binary vector $\bb \in \{-1,1\}^L$, where $L \ll D$. 

Many hashing methods have been proposed in the literature. They can be divided into two categories: data-independent and data-dependent. Methods in data-independent category~\cite{lsh_vldb09,KLSH_iccv09,KLSH_nips09,DBLP:journals/pami/KulisJG09} rely on random projections for constructing hash functions. Methods in data-dependent category use the available training data to learn the hash functions in unsupervised~\cite{DBLP:conf/nips/WeissTF08,DBLP:conf/cvpr/GongL11,DBLP:conf/cvpr/HeWS13,CVPR12:SphericalHashing,conf/nips/KongL12} or supervised manner~\cite{DBLP:journals/pami/StrechaBBF12,CVPR12:Hashing,DBLP:conf/nips/0002FS12,CVPR2014Lin,Kulis_learningto,Shen_2015_CVPR}. The review of data-independent/data-dependent hashing methods can be found in recent surveys~\cite{DBLP:journals/corr/WangLKC15,DBLP:journals/corr/WangSSJ14,Grauman_review}.


One difficult problem in hashing is to deal with the binary constraint on the codes.  Specifically, the outputs of the hash functions have to be binary. In general, this binary constraint leads to a NP-hard mixed-integer optimization problem. To handle this difficulty, most aforementioned methods relax the constraint during the learning of hash functions. With this relaxation, the continuous codes are learned first.  Then, the codes are binarized (e.g., with thresholding). This relaxation greatly simplifies the original binary constrained problem.  However,  the solution can be suboptimal, i.e., the binary codes resulting from thresholded continuous codes could be inferior to those that are obtained by including the binary constraint in the learning.

Furthermore, a good hashing method should produce binary codes with these properties~\cite{DBLP:conf/nips/WeissTF08}: (i) similarity preserving, i.e., (dis)similar inputs should likely have (dis)similar binary codes; (ii) independence, i.e.,  different bits in the binary codes are independent to each other; (iii) balance, i.e.,  each bit has a $50\%$ chance of being $1$ or $-1$.
Note that direct incorporation of the independent and balance properties can complicate the learning. Previous work has used some relaxation or approximation to work around the problem~\cite{DBLP:conf/cvpr/GongL11,Liong_2015_CVPR,DBLP:journals/pami/WangKC12}, but there may be some performance degradation.

\vspace{-0.3cm}
\subsection{Related work} 
Our work is inspired by a few recent successful hashing methods which define hash functions as a neural network~\cite{DBLP:SeH,Liong_2015_CVPR,BA_CVPR15}. 
We propose an improved design to address their limitations.
In Semantic Hashing~\cite{DBLP:SeH}, the model is formed by a stack of Restricted Boltzmann Machine, and a pretraining step is required. This model does not consider the independence and balance of the codes. 
In Binary Autoencoder~\cite{BA_CVPR15}, a linear autoencoder is used as hash functions. As this model only uses one hidden layer, it may not well capture the information of inputs.
Extending~\cite{BA_CVPR15} with multiple, nonlinear layers is not straight-forward because of the binary constraint. They also do not consider the independence and balance of codes. In Deep Hashing~\cite{Liong_2015_CVPR}, a deep neural network is used as hash functions. However, this model does not fully take into account the similarity preserving.
They also apply some relaxation in arriving the independence and balance of codes and this may degrade the performance.

In order to handle the binary constraint, Semantic Hashing~\cite{DBLP:SeH} first solves the relaxed problem by discarding the constraint and then thresholds the solved continuous solution. In Deep Hashing (DH)~\cite{Liong_2015_CVPR}, the output of the last layer, $\H^{n}$, is binarized by the $sgn$ function.  They include a term in the objective function to reduce this binarization loss: $\left(sgn(\H^{n}) - \H^{n} \right)$. Solving the objective function of DH~\cite{Liong_2015_CVPR} is difficult because the $sgn$ function is non-differentiable. 
The authors in~\cite{Liong_2015_CVPR} work  around this difficulty by assuming that the $sgn$ function is differentiable everywhere.  In Binary Autoencoder (BA)~\cite{BA_CVPR15},  the outputs of the hidden layer are passed into a step function to binarize the codes.  Incorporating the step function in the learning leads to a non-smooth objective function and the optimization is NP-complete.  To handle this difficulty, they use binary SVMs to learn the model parameters in the case when there is only a single hidden layer.

\vspace{-0.3cm}
\subsection{Contribution}
In this work, we first propose a novel deep network model and learning algorithm for unsupervised hashing. In order to achieve binary codes, instead of involving the $sgn$ or step function as in~\cite{Liong_2015_CVPR,BA_CVPR15}, our proposed network design constrains one layer to directly output the binary codes (therefore, the network is named as \textit{Binary Deep Neural Network}). 
Moreover, we propose to directly incorporate the independence and balance properties without relaxing them. Furthermore, we include the similarity preserving in our objective function. 
The resulting optimization with these binary and direct constraints is NP-hard.  
We propose to attack this challenging problem with alternating optimization and careful relaxation. 
To enhance the discriminative power of the binary codes, we then extend our method to supervised hashing by leveraging the label information such that the binary codes preserve the semantic similarity between samples. The solid experiments on three benchmark datasets show the improvement of the proposed methods over state-of-the-art hashing methods. 

The remaining of this paper is organized as follows. Section~\ref{sec:UH-BDNN} and Section~\ref{sec:eva_UH-BDNN} present and evaluate the proposed unsupervised hashing method, respectively. Section~\ref{sec:SH-BDNN} and Section~\ref{sec:eva_SH-BDNN} present and evaluate the proposed supervised hashing method, respectively. Section~\ref{sec:conclusion} concludes the paper.
\vspace{-0.3cm}
\section{Unsupervised Hashing with Binary Deep Neural Network (UH-BDNN)}
\label{sec:UH-BDNN}
\subsection{Formulation of UH-BDNN}
\label{subsec:formular_un}
We summarize the notations in Table~\ref{tab:notation}.
\begin{table}[!t]
\footnotesize
\centering
\caption{Notations and their corresponding meanings.}
\label{tab:notation}
\begin{tabular}{|l|l|}
\hline
Notation	&Meaning \\ \hline
$\X$ &$\X = \{\x_i\}_{i=1}^{m} \in \R^{D\times m}$: set of $m$ training samples; \\
	 &each column of $\X$ corresponds to one sample\\ \hline	
$\B$ &$\B = \{\bb_i\}_{i=1}^{m} \in \{-1,+1\}^{L\times m}$: binary code of $\X$ \\ \hline
$L$  &Number of bits in the output binary code to encode a sample \\ \hline
$n$  &Number of layers (including input and output layers) \\ \hline
$s_l$&Number of units in layer $l$	\\ \hline
$f^{(l)}$ &Activation function of layer $l$ \\ \hline		
$\W^{(l)}$&$\W^{(l)} \in \R^{s_{l+1}\times s_l}$: weight matrix connecting layer $l+1$  and layer $l$	\\ \hline
$\cc^{(l)}$&$\cc^{(l)} \in \R^{s_{l+1}}$:bias vector for units in layer $l+1$ \\ \hline
$\H^{(l)}$ &$\H^{(l)} = f^{(l)}\left(\W^{(l-1)}\H^{(l-1)} + \cc^{(l-1)}\1_{1\times m}\right)$: output values of layer $l$; \\ 
	      &convention: $\H^{(1)} = \X$ \\ \hline
$\1_{a\times b}$ & Matrix has $a$ rows, $b$ columns and all elements equal to 1	\\ \hline
\end{tabular}
\end{table}
\begin{figure}[!t]
\centering
\includegraphics[scale=0.32]{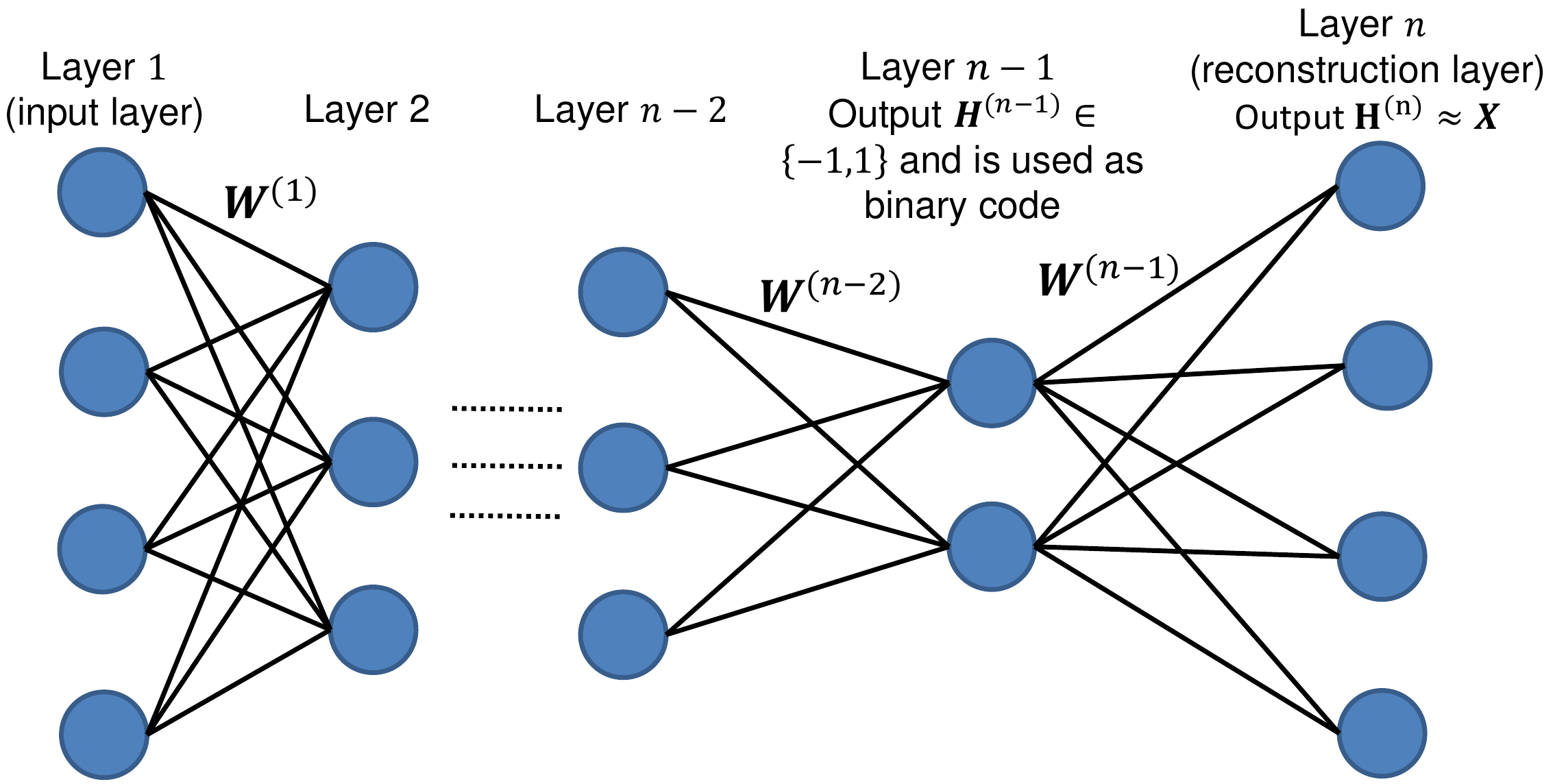}
\caption{The illustration of our network ($D=4,L=2$). In our proposed network design, the outputs of layer $n-1$ are constrained to $\{-1,1\}$ and are used as the binary codes. During training, these codes are used to reconstruct the input samples at the final layer.}
\label{fig:network}
\end{figure}
In our work, the hash functions are defined by a deep neural network. In our proposed design, we use different activation functions in different layers.  Specifically, we use the sigmoid function as activation function for layers $2,\cdots,n-2$, and the identity function as activation function for layer $n-1$ and layer $n$. 
Our idea is to learn the network such that the output values of the \textit{penultimate layer} (layer $n-1$)
can be used as the binary codes.  We introduce constraints in the learning algorithm such that the output values at the layer $n-1$ 
have the following desirable properties:  (i) belonging to $\{-1,1\}$; (ii) similarity preserving; (iii) independent and (iv) balancing. Figure~\ref{fig:network} illustrates our  network for the case $D=4,L=2$.

Let us start with first two properties of the codes, i.e., belonging to $\{-1,1\}$ and similarity preserving. To achieve the binary codes having these two properties, we propose to optimize the following constrained objective function
\vspace{-0.2cm}\footnotesize
\begin{eqnarray}
\min_{\W,\cc} J &=& \frac{1}{2m} \norm{\X-\left(\W^{(n-1)}\H^{(n-1)}+\cc^{(n-1)}\1_{1\times m}\right)}^2 +\frac{\lambda_1}{2}\sum_{l=1}^{n-1} \norm{\W^{(l)}}^2 \label{eq:obj_ori}
\end{eqnarray}
\begin{equation}
\hspace{-2cm}\textrm{s.t. } \H^{(n-1)} \in \{-1,1\}^{L\times m} \label{eq:binary0}
\end{equation} 
\normalsize 
The constraint~(\ref{eq:binary0}) is to ensure the first property. As the activation function for the last layer is the identity function, the term $\left(\W^{(n-1)}\H^{(n-1)}+\cc^{(n-1)}\1_{1\times m}\right)$ is the output of the last layer. The first term of~(\ref{eq:obj_ori}) makes sure that the binary code gives a good reconstruction of $\X$. 
It is worth noting that the reconstruction criterion has been used as an indirect way for preserving the similarity in state-of-the-art unsupervised hashing methods~\cite{DBLP:conf/cvpr/GongL11,BA_CVPR15,DBLP:SeH}, i.e., it encourages (dis)similar inputs map to (dis)similar binary codes. 
The second term is a regularization that tends to decrease the magnitude of the weights, and this helps to prevent overfitting. 
Note that in our proposed design, we constrain to directly output the binary codes at one layer, and this avoids the difficulties with the $sgn$/step function such as non-differentiability.  On the other hand, our formulation with (\ref{eq:obj_ori}) under the binary constraint~(\ref{eq:binary0}) is very difficult to solve. It is a mixed-integer problem which is NP-hard. We propose to attack the problem using alternating optimization by introducing an auxiliary variable. Using the auxiliary variable $\B$, we reformulate the objective function~(\ref{eq:obj_ori}) under constraint~(\ref{eq:binary0}) as                    

\vspace{-0.3cm} \footnotesize 
\begin{eqnarray}
\min_{\W,\cc,\B} J &=& \frac{1}{2m} \norm{\X-\W^{(n-1)}\B-\cc^{(n-1)}\1_{1\times m}}^2 +\frac{\lambda_1}{2}\sum_{l=1}^{n-1} \norm{\W^{(l)}}^2 \label{eq:obj_2}
\end{eqnarray}
\begin{equation}
\textrm{s.t. }\B = \H^{(n-1)} \label{eq:binary}
\end{equation} 
\vspace{-0.3cm}
\begin{equation}
\hspace{3.7em}\B \in \{-1,1\}^{L\times m} \label{eq:binary_add}
\end{equation} 
\normalsize
The benefit of introducing the auxiliary variable $\B$ is that we can decompose the difficult constrained optimization problem~(\ref{eq:obj_ori}) into two  sub-optimization problems. Then, we can iteratively solve the optimization by using alternating optimization with respect to $(\W,\cc)$ and $\B$ while holding the other fixed. We will discuss the details of the alternating optimization in a moment.
Using the idea of the quadratic penalty method~\cite{Nocedal06}, we relax the equality constraint (\ref{eq:binary}) by solving the following constrained objective function 

\vspace{-0.3cm}\footnotesize
\begin{eqnarray}
\min_{\W,\cc,\B} J &=& \frac{1}{2m} \norm{\X-\W^{(n-1)}\B-\cc^{(n-1)}\1_{1\times m}}^2 \nonumber \\ 
{}&&\hspace{0em}+\frac{\lambda_1}{2}\sum_{l=1}^{n-1} \norm{\W^{(l)}}^2 + \frac{\lambda_2}{2m}\norm{\H^{(n-1)}-\B}^2  \label{eq:obj_4}
\end{eqnarray}
\begin{equation}
\hspace{-1.5cm}\textrm{s.t. }\B \in \{-1,1\}^{L\times m} \label{eq:binary_H}
\end{equation}
\normalsize
The third term in (\ref{eq:obj_4}) measures the (equality) constraint violation. By setting the penalty parameter $\lambda_2$ sufficiently large, we penalize the constraint violation severely, thereby forcing the minimizer of the penalty function (\ref{eq:obj_4}) closer to the feasible region of the original constrained function (\ref{eq:obj_2}). 

Now let us consider the two remaining properties of the codes, i.e., independence and balance. 
Unlike previous works which use some relaxation or approximation on the independence and balance properties~\cite{DBLP:conf/cvpr/GongL11,Liong_2015_CVPR,DBLP:journals/pami/WangKC12}, we propose to encode these properties strictly and directly based on the binary outputs of our layer $n-1$ \footnote{Alternatively, we can constrain the independence and balance on $\B$. This, however, makes the optimization very difficult.}. Specifically, we encode the independence and balance properties of the codes by having the fourth and the fifth term respectively in the following constrained objective function

\vspace{-0.3cm}\footnotesize
\begin{eqnarray}
\min_{\W,\cc,\B} J &=& \frac{1}{2m} \norm{\X-\W^{(n-1)}\B-\cc^{(n-1)}\1_{1\times m}}^2 +\frac{\lambda_1}{2}\sum_{l=1}^{n-1} \norm{\W^{(l)}}^2\nonumber \\ 
{}&&\hspace{-5em} + \frac{\lambda_2}{2m}\norm{\H^{(n-1)}-\B}^2 +\frac{\lambda_3}{2}\norm{\frac{1}{m}\H^{(n-1)}(\H^{(n-1)})^T-\I}^2 +\frac{\lambda_4}{2m}\norm{\H^{(n-1)}\1_{m\times 1}}^2 \label{eq:obj_5}
\end{eqnarray}
\begin{equation}
\hspace{-2cm}\textrm{s.t. }\B \in \{-1,1\}^{L\times m} \label{eq:binary_H1}
\end{equation}
\normalsize
(\ref{eq:obj_5}) under constraint (\ref{eq:binary_H1}) is our final formulation.  Before discussing  how to solve it, let us present the differences between our work and the recent deep learning based-hashing models Deep Hashing~\cite{Liong_2015_CVPR} and Binary Autoencoder~\cite{BA_CVPR15}.

The first important difference between our model and  Deep Hashing~\cite{Liong_2015_CVPR} / Binary Autoencoder~\cite{BA_CVPR15} is  the way to achieve the binary codes. Instead of involving the $sgn$ or step function as in~\cite{Liong_2015_CVPR,BA_CVPR15}, 
we constrain the network to directly output the binary codes at one layer. 
Other differences are presented as follows.
\vspace{-0.3cm}
\paragraph{Comparison to Deep Hashing (DH)~\cite{Liong_2015_CVPR}:} the deep model of DH is learned by the following formulation:

\vspace{-0.4cm}\footnotesize
\begin{eqnarray}
\min_{\W,\cc} J &=& \frac{1}{2} \norm{sgn(\H^{(n)}) - \H^{(n)}}^2 - \frac{\alpha_1}{2m}tr\left(\H^{(n)}(\H^{(n)})^T \right)
\nonumber \\ 
{}&&\hspace{-6em}+\frac{\alpha_2}{2}\sum_{l=1}^{n-1} \norm{\W^{(l)}(\W^{(l)})^T - \I}^2 
+\frac{\alpha_3}{2}\sum_{l=1}^{n-1} \left(\norm{\W^{(l)}}^2 + \norm{\cc^{(l)}}^2 \right)  
\nonumber \label{eq:obj_dh}
\end{eqnarray}
\normalsize
The DH's model does not have the reconstruction layer. They apply $sgn$ function to the outputs at the top layer of the network to obtain the binary codes.
The first term aims to minimize quantization loss
when applying the $sgn$ function to the outputs at the top layer.
The balancing and the independent properties are contained in the second and the third terms~\cite{Liong_2015_CVPR}. 
It is worth noting that minimizing DH's objective function is difficult due to the non-differentiable of \textit{sgn} function. The authors work around this difficulty by assuming that \textit{sgn} function is differentiable everywhere. 

Contrary to DH, we propose a different model design.  In particular, our model encourages the similarity preserving by having the reconstruction layer in the network. For the balancing property, they maximize $tr \left(\H^{(n)}(\H^{(n)})^T \right)$. According to~\cite{DBLP:journals/pami/WangKC12}, maximizing this term is only an approximation in arriving the balancing property. In our objective function, the balancing property is directly enforced on the codes by the term $\norm{\H^{(n-1)}\1_{m\times 1}}^2$. 
For the independent property, DH uses a relaxed orthogonality constraint $\norm{\W^{(l)}(\W^{(l)})^T - \I}^2$, i.e., constraining on the network weights $\W$.  On the contrary, we (once again) directly constrain on the codes using $\norm{\frac{1}{m}\H^{(n-1)}(\H^{(n-1)})^T-\I}^2$.
Incorporating the strict constraints can lead to better performance.

\vspace{-0.3cm}
\paragraph{Comparison to Binary Autoencoder (BA)~\cite{BA_CVPR15}:} the differences between our model and BA are quite clear. 
BA as described in~\cite{BA_CVPR15} is a shallow linear autoencoder network with one hidden layer. The BA's hash function is a linear transformation of the input followed by the step function to obtain the binary codes.
In BA, by treating the encoder layer as binary classifiers, they use binary SVMs 
to learn the weights of the linear transformation.  On the contrary, our hash function is defined by multiple,  hierarchical layers of nonlinear and linear transformations. 
It is not clear if the binary SVMs approach in BA can be used to learn the weights in our deep architecture with multiple layers. Instead, we use alternating optimization to derive a backpropagation algorithm to learn the weights in all layers.  
Another difference is that our model ensures the independence and balance of the binary codes while BA does not.
Note that independence and balance properties may not be easily incorporated in their framework, as these would complicate their objective function and the optimization problem may become very difficult to solve.

\vspace{-0.3cm}
\subsection{Optimization}
\label{subsec:UDH_opt}
In order to solve~(\ref{eq:obj_5}) under constraint~(\ref{eq:binary_H1}), we propose to use  alternating optimization over $(\W,\cc)$ and $\B$.
\vspace{-0.3cm}
\subsubsection{$(\W,\cc)$ step}
\label{subsub:W_step_un}
When fixing $\B$, the problem becomes unconstrained optimization. We use \textit{L-BFGS}~\cite{Liu89onthe} optimizer with backpropagation for solving. The gradient of the objective function $J$ (\ref{eq:obj_5}) w.r.t. different parameters are computed as follows.

At $l = n-1$, we have

\vspace{-0.2cm}\footnotesize
\begin{equation}
\frac{\partial J}{\partial \W^{(n-1)}} = \frac{-1}{m}(\X-\W^{(n-1)}\B-\cc^{(n-1)}\1_{1\times m})\B^T + \lambda_1 \W^{(n-1)}
\end{equation}
\begin{equation}
\frac{\partial J}{\partial \cc^{(n-1)}} = \frac{-1}{m}\left( (\X-\W^{(n-1)}\B)\1_{m\times 1}-m\cc^{(n-1)} \right)
\end{equation}
\normalsize
For other layers, let us define

\vspace{-0.4cm}\footnotesize
\begin{eqnarray}
\Delta^{(n-1)} &=& \left[ \frac{\lambda_2}{m}\left( \H^{(n-1)}-\B \right)+\frac{2\lambda_3}{m}\left( \frac{1}{m}\H^{(n-1)}(\H^{(n-1)})^T - \I\right)\H^{(n-1)} \right. \nonumber \\
 {}&& \left. +\frac{\lambda_4}{m}\left( \H^{(n-1)}\1_{m\times m} \right) \right]\odot f^{(n-1)'}(\Z^{(n-1)})
\end{eqnarray}
\begin{equation}
\Delta^{(l)} = \left( (\W^{(l)})^T\Delta^{(l+1)} \right) \odot f^{(l)'}(\Z^{(l)}),\forall l = n-2,\cdots,2
\end{equation}
\normalsize
where $\odot$ denotes Hadamard product; \small$\Z^{(l)} = \W^{(l-1)}\H^{(l-1)} + \cc^{(l-1)}\1_{1\times m}$, $l=2,\cdots,n$.\normalsize

Then, $\forall l = n-2,\cdots,1$, we have

\footnotesize
\begin{equation}
\frac{\partial J}{\partial \W^{(l)}} = \Delta^{(l+1)}(\H^{(l)})^T +\lambda_1\W^{(l)}
\end{equation}
\begin{equation}
\frac{\partial J}{\partial \cc^{(l)}} = \Delta^{(l+1)}\1_{m\times 1}
\end{equation}
\normalsize
\vspace{-0.4cm}
\subsubsection{$\B$ step}
\label{subsub:B_step_un}
When fixing $(\W,\cc)$, we can rewrite problem~(\ref{eq:obj_5}) as 

\vspace{-0.4cm}\footnotesize
\begin{eqnarray}
\min_{\B} J &=& \norm{\X-\W^{(n-1)}\B-\cc^{(n-1)}\1_{1\times m}}^2 +\lambda_2 \norm{\H^{(n-1)}-\B}^2
\label{eq:B_step_un}
\end{eqnarray} 
\begin{equation}
\hspace{-1cm}\textrm{s.t. }\B \in \{-1,1\}^{L\times m} \label{eq:binary_H11}
\end{equation}
\normalsize
We adaptively use the recent method \textit{discrete cyclic coordinate descent}~\cite{Shen_2015_CVPR} to  iteratively solve  $\B$, i.e., row by row. The advantage of this method is that if we fix $L-1$ rows of $\B$ and only solve for the remaining row, we can achieve a closed-form solution for that row. 

Let $\V = \X-\cc^{(n-1)}\1_{1\times m}$; $\Q = (\W^{(n-1)})^T\V+\lambda_2\H^{(n-1)}$. For $k=1,\cdots L$, let $\w_k$ be $k^{th}$ column of $\W^{(n-1)}$; $\W_1$ be matrix $\W^{(n-1)}$ excluding $\w_k$; $\q_k$ be $k^{th}$ column of $\Q^T$; $\bb_k^T$ be $k^{th}$ row of $\B$; $\B_1$ be matrix of $\B$ excluding $\bb_k^T$. We have closed-form for $\bb_k^T$ as
\footnotesize
\begin{equation}
\bb_k^T = sgn(\q^T - \w_k^T\W_1\B_1)
\end{equation}
\normalsize
The proposed UH-BDNN method is summarized in Algorithm~\ref{alg1}. In the Algorithm~\ref{alg1}, $\B_{(t)}$ and $(\W,\cc)_{(t)}$ are values of $\B$ and $\{\W^{(l)},\cc^{(l)}\}_{l=1}^{n-1}$ at iteration $t$.
\begin{algorithm}[!t]
	\scriptsize
	\caption{Unsupervised Hashing with Binary Deep Neural Network (UH-BDNN)}
	\begin{algorithmic}[1] 
		\Require 
			\Statex $\X = \{\x_i\}_{i=1}^{m} \in \R^{D\times m}$: training data; $L$: code length; $T$: maximum iteration number; $n$: number of layers; $\{s_l\}_{l=2}^{n}$: number of units of layers $2 \to n$ (note: $s_{n-1} = L$, $s_n = D$); $\lambda_1, \lambda_2, \lambda_3, \lambda_4$.
		\Ensure 
			\Statex 
			Parameters $\{\W^{(l)},\cc^{(l)}\}_{l=1}^{n-1}$
			\Statex 
			\State Initialize $\B_{(0)} \in \{-1,1\}^{L\times m}$ using ITQ~\cite{DBLP:conf/cvpr/GongL11}
			\State Initialize $\{\cc^{(l)}\}_{l=1}^{n-1} = \mathbf{0}_{s_{l+1}\times 1}$. Initialize $\{\W^{(l)}\}_{l=1}^{n-2}$ by getting the top $s_{l+1}$ eigenvectors from the covariance matrix of $\H^{(l)}$. Initialize $\W^{(n-1)} = \I_{D\times L}$
			\State Fix $\B_{(0)}$, compute $(\W,\cc)_{(0)}$ with $(\W,\cc)$ step using initialized $\{\W^{(l)},\cc^{(l)}\}_{l=1}^{n-1}$ (line 2) as starting point for L-BFGS.
			\For{$t = 1 \to T$}
				\State Fix $(\W,\cc)_{(t-1)}$, compute $\B_{(t)}$ with $\B$ step
				\State Fix $\B_{(t)}$, compute $(\W,\cc)_{(t)}$ with $(\W, \cc)$ step using $(\W,\cc)_{(t-1)}$ as starting point for L-BFGS.
			\EndFor
			\State Return 
			$(\W,\cc)_{(T)}$
    \end{algorithmic}
    \label{alg1}
\end{algorithm}
\vspace{-0.3cm}
\section{Evaluation of Unsupervised Hashing with Binary Deep Neural Network (UH-BDNN)}
\label{sec:eva_UH-BDNN}
This section evaluates the proposed UH-BDNN and compares it with following state-of-the-art unsupervised hashing methods: Spectral Hashing (SH)~\cite{DBLP:conf/nips/WeissTF08},  Iterative Quantization (ITQ)~\cite{DBLP:conf/cvpr/GongL11}, Binary Autoencoder (BA)~\cite{BA_CVPR15}, Spherical Hashing (SPH)~\cite{CVPR12:SphericalHashing}, K-means Hashing (KMH)~\cite{DBLP:conf/cvpr/HeWS13}. For all compared methods, we use the implementations and the suggested parameters provided by the authors.
\subsection{Dataset, evaluation protocol, and implementation note}
\label{subsec:data-imp-eva}
\subsubsection{Dataset}
{CIFAR10~\cite{Krizhevsky09}} dataset consists of 60,000 images of 10 classes. The training set (also used as database for retrieval) contains 50,000 images. The query set contains 10,000 images. 
Each image is represented by a 800-dimensional feature vector extracted by PCA from
4096-dimensional CNN feature produced by AlexNet~\cite{jia2014caffe}. 

{MNIST~\cite{mnistlecun}} dataset  consists of 70,000 handwritten digit
images of 10 classes. The training set (also used as database for retrieval) contains 60,000 images. The query set contains 10,000 images. Each image is represented by a 784 dimensional gray-scale feature vector by using its intensity.

{SIFT1M~\cite{herve_pami2011}} dataset contains 128 dimensional SIFT vectors~\cite{SIFT_Lowe}. There are 1M vectors used as database for retrieval; 100K vectors for training (separated from retrieval database) and 10K vectors for query.  

\vspace{-0.4cm}
\subsubsection{Evaluation protocol} 
We follow the standard setting in unsupervised hashing~\cite{DBLP:conf/cvpr/GongL11,CVPR12:SphericalHashing,DBLP:conf/cvpr/HeWS13,BA_CVPR15} using Euclidean nearest neighbors as the  ground truths for queries. Number of ground truths are set as in~\cite{BA_CVPR15}, i.e., for CIFAR10 and MNIST datasets, for each query, we use $50$ its Euclidean nearest neighbors as ground truths; for large scale dataset SIFT1M, for each query, we use $10,000$ its Euclidean nearest neighbors as ground truths. 
We use the following evaluation metrics which have been used in state of the art~\cite{DBLP:conf/cvpr/GongL11,BA_CVPR15,Liong_2015_CVPR} to measure the performance of methods. 1) mean Average Precision (mAP); 2) precision of Hamming radius $2$ (precision$@2$) which measures precision on retrieved images having Hamming distance to query $\le 2$ (if no images satisfy, we report zero precision). Note that as computing mAP is slow on large dataset SIFT1M, we consider top $10,000$ returned neighbors when computing mAP.

\vspace{-0.4cm}
\subsubsection{Implementation note}
In our deep model, we use $n=5$ layers. 
The parameters $\lambda_1$, $\lambda_2$, $\lambda_3$ and $\lambda_4$ are empirically set by cross validation 
as $10^{-5}$, $5\times 10^{-2}$, $10^{-2}$ and $10^{-6}$, respectively. The max iteration number $T$ is empirically set to 10. 
The number of units in hidden layers $2,3,4$ are empirically set as $[90 \to 20 \to 8]$, $[90 \to 30 \to 16]$, $[100 \to 40 \to 24]$ and $[120 \to 50 \to 32]$ for the 8, 16, 24 and 32 bits, respectively. 
\begin{figure*}[!t]
\centering
\subfigure[CIFAR10]{
       \includegraphics[scale=0.28]{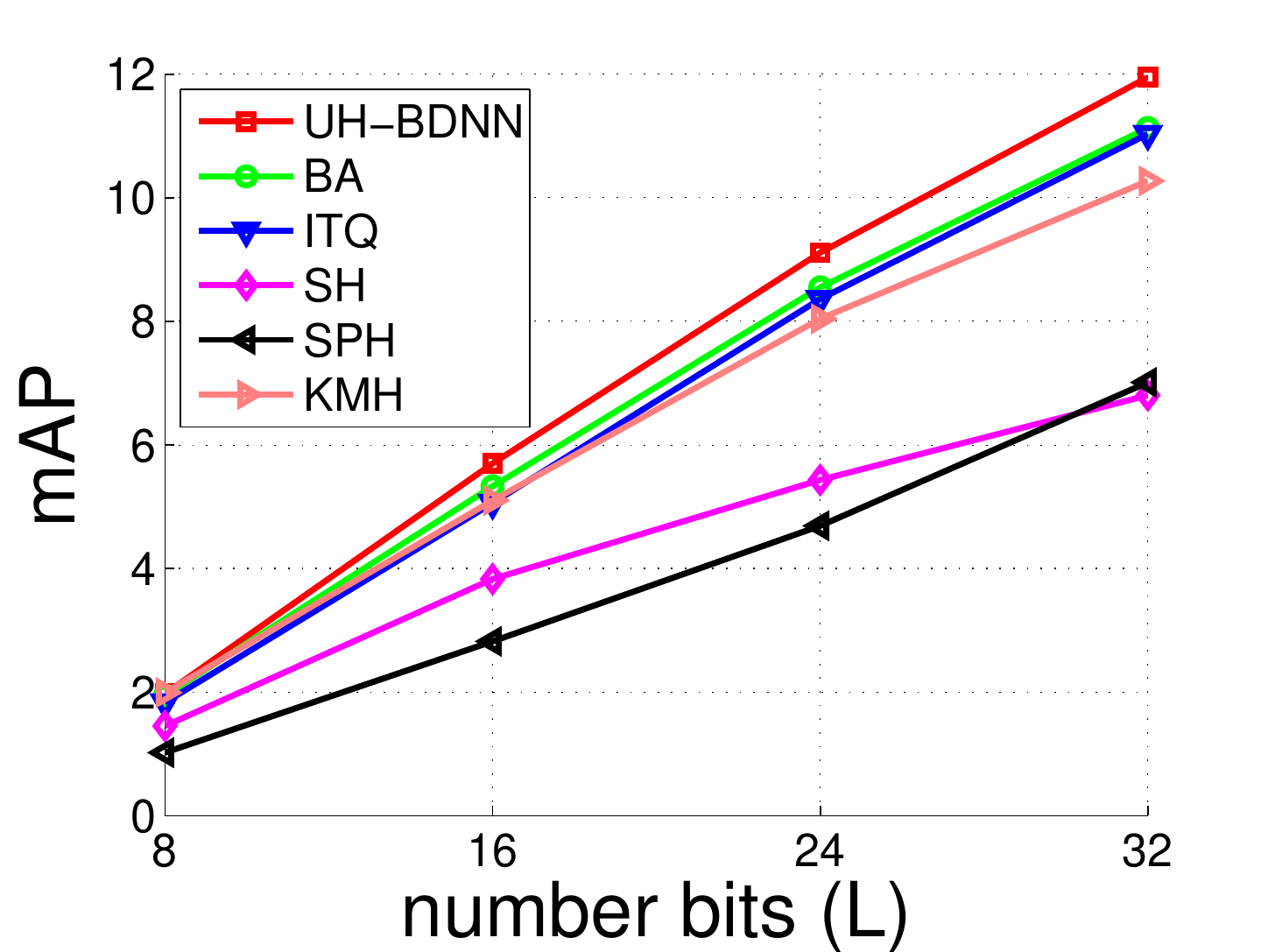}
       \label{fig:cifar_mAP}
}\hspace{-0.5cm}
\subfigure[MNIST]{
       \includegraphics[scale=0.28]{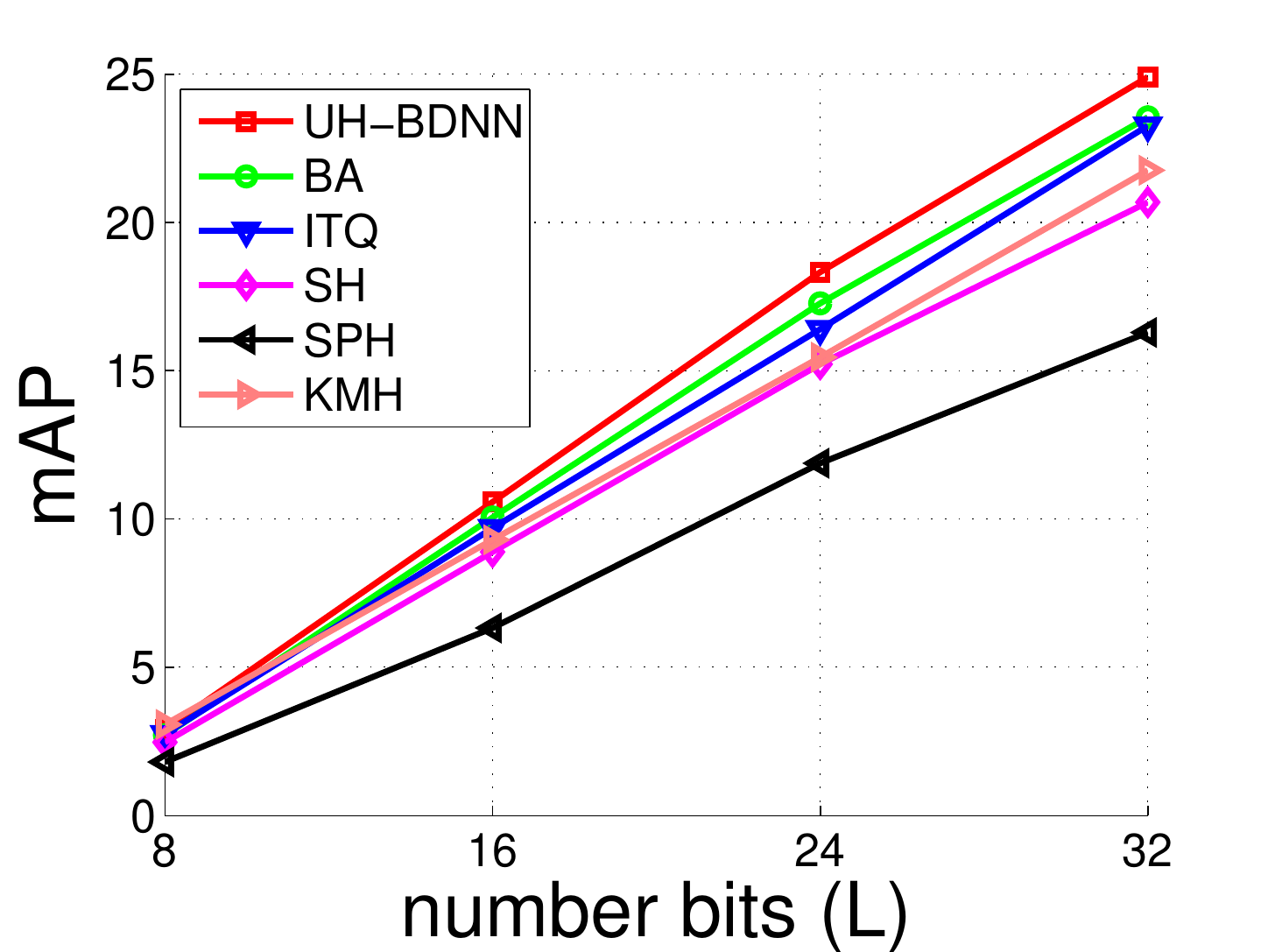} 
       \label{fig:mnist_mAP}
}\hspace{-0.5cm}
\subfigure[SIFT1M]{
       \includegraphics[scale=0.28]{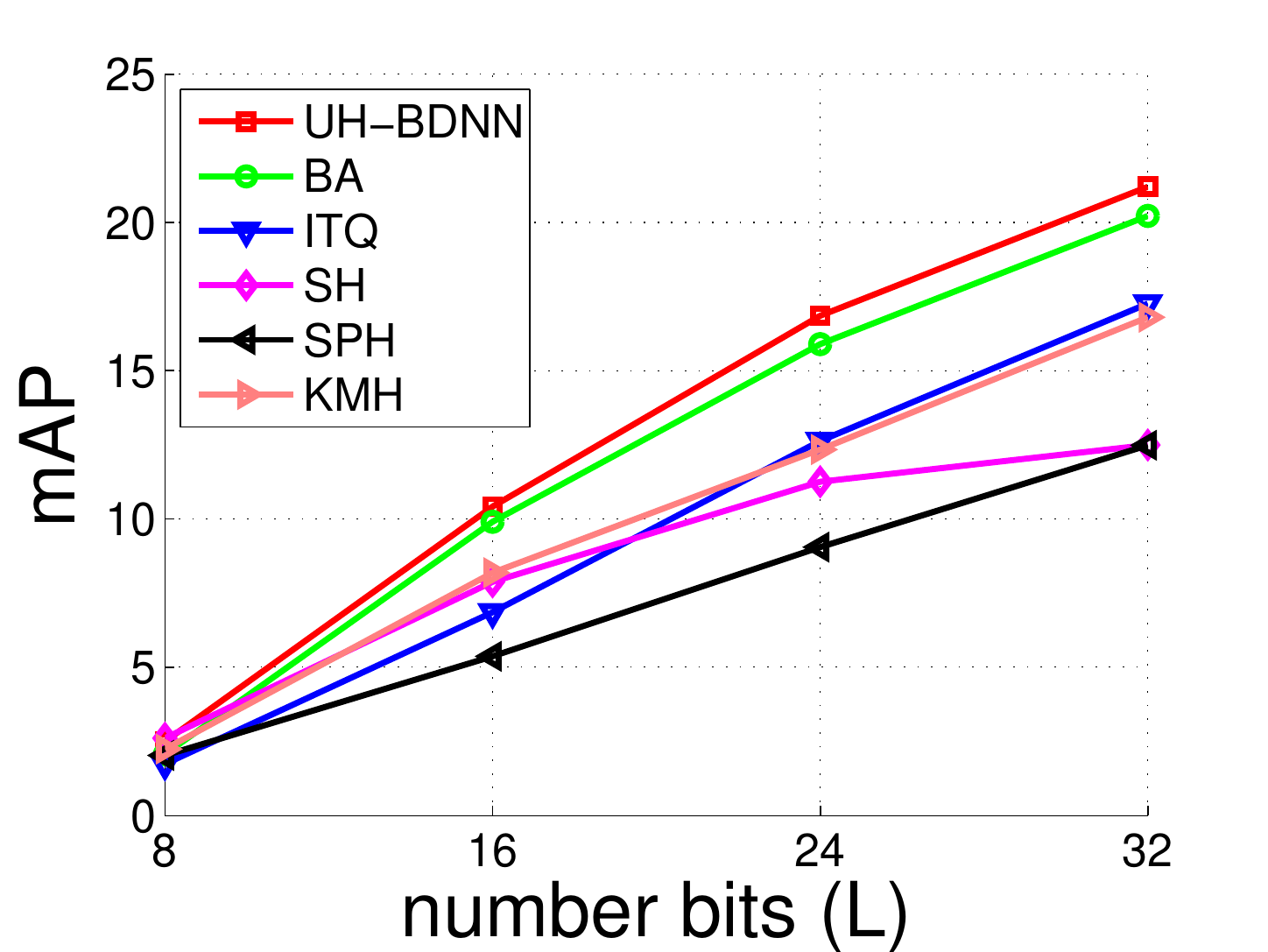} 
       \label{fig:sift1m_mAP}
}
\caption[]{mAP comparison between UH-BDNN and state-of-the-art unsupervised hashing methods on CIFAR10, MNIST, and SIFT1M.}
\label{fig:mAP_cifar10_mnist_sift1m}
\end{figure*}
\begin{table}[!t]
   \centering
   \footnotesize
   \caption{Precision at Hamming distance $r=2$ comparison between UH-BDNN and state-of-the-art unsupervised hashing methods on CIFAR10, MNIST, and SIFT1M.}
    \begin{tabular}{|l|c| c| c| c|c| c| c| c|c| c| c| c|}
		\hline
	  \multirow{2}{*}{} & \multicolumn{4}{|c|}{CIFAR10} & \multicolumn{4}{|c|}{MNIST} & \multicolumn{4}{|c|}{SIFT1M}\\
\cline{1-13}	$L$    &8 &16 &24 &32    &8 &16 &24 &32  &8 &16 &24 &32   \\ \hline 
UH-BDNN	   						 &0.55 &5.79 &22.14 &18.35   &0.53 &6.80 &29.38 &38.50 
    &4.80 &25.20 &62.20 &80.55 \\ \hline
BA\cite{BA_CVPR15}   			 &0.55 &5.65 &20.23 &17.00   &0.51 &6.44 &27.65 &35.29
	&3.85 &23.19 &61.35 &77.15 \\ \hline
ITQ\cite{DBLP:conf/cvpr/GongL11} &0.54 &5.05 &18.82 &17.76   &0.51 &5.87 &23.92 &36.35
	&3.19 &14.07 &35.80 &58.69 \\ \hline
SH\cite{DBLP:conf/nips/WeissTF08}&0.39 &4.23 &14.60 &15.22   &0.43 &6.50 &27.08 &36.69
    &4.67 &24.82 &60.25 &72.40 \\ \hline
SPH\cite{CVPR12:SphericalHashing}&0.43 &3.45 &13.47 &13.67   &0.44 &5.02 &22.24 &30.80
	&4.25 &20.98 &47.09 &66.42 \\ \hline
KMH\cite{DBLP:conf/cvpr/HeWS13}	 &0.53 &5.49 &19.55 &15.90   &0.50 &6.36 &25.68 &36.24 			
	&3.74 &20.74 &48.86 &76.04 \\ \hline
	  \end{tabular}
	  \label{tab:soa-UNsup-cifar10-mnist-sift1m-pre}
	  \vspace{-0.2cm}
\end{table}
\vspace{-0.3cm}
\subsection{Retrieval results}
\vspace{-0.2cm}
Fig.~\ref{fig:mAP_cifar10_mnist_sift1m} and Table~\ref{tab:soa-UNsup-cifar10-mnist-sift1m-pre} show comparative mAP and precision of Hamming radius $2$ (precision$@2$), respectively. We find the following observations are consistent for all three datasets. In term of mAP, the proposed UH-BDNN comparable or outperforms other methods at all code lengths. The improvement is more clear at high code length, i.e., $L=24,32$. The mAP of UH-BDNN consistently outperforms that of binary autoencoder (BA)~\cite{BA_CVPR15}, which is the current state-of-the-art unsupervised hashing method. In term of precision$@2$, UH-BDNN is comparable to other methods at low $L$, i.e., $L = 8, 16$. At $L = 24, 32$, UH-BDNN significantly outperforms other methods. 
\vspace{-0.3cm}
\paragraph{Comparison with Deep Hashing (DH)~\cite{Liong_2015_CVPR}}
As the implementation of DH is not available, we set up the experiments on CIFAR10 and MNIST similar to~\cite{Liong_2015_CVPR} to make a fair comparison. For each dataset, we randomly sample 1,000 images, 100 per class, as query set; the remaining images are used as training/database set. Follow~\cite{Liong_2015_CVPR}, for CIFAR10, each image is represented by 512-$D$ GIST descriptor~\cite{gist}. 
The ground truths of queries are based on their class labels. 
Similar to~\cite{Liong_2015_CVPR}, we report comparative results in term of mAP and the precision of Hamming radius $r=2$. 
The comparative results are presented in the Table~\ref{tab:compare_DH_cifar512}.
\begin{table}[!t]
\vspace{-0.2em}
\footnotesize
\centering
\caption{Comparison with Deep Hashing (DH)~\cite{Liong_2015_CVPR}. The results of DH are  cited from~\cite{Liong_2015_CVPR}.} 
\label{tab:compare_DH_cifar512}
\begin{tabular}{|c|c c|c c|c c|c c|}
\hline
		&\multicolumn{4}{|c|}{CIFAR10} 	 &\multicolumn{4}{|c|}{MNIST}\\  \hline
	&\multicolumn{2}{|c|}{mAP} 	 &\multicolumn{2}{|c|}{precision$@2$} &\multicolumn{2}{|c|}{mAP} 	 &\multicolumn{2}{|c|}{precision$@2$}\\ 
	L						&16	   &32	  &16	 &32    &16    &32    &16    &32 \\\hline
DH~\cite{Liong_2015_CVPR}   &16.17 &16.62 &23.33 &15.77 &43.14 &44.97 &66.10 &73.29\\
UH-BDNN	                    &17.83 &18.52 &24.97 &18.85 &45.38 &47.21 &69.13 &75.26\\\hline 
\end{tabular}
\vspace{-0.3cm}
\end{table} 
It is clearly showed in Table~\ref{tab:compare_DH_cifar512} that the proposed UH-BDNN outperforms DH~\cite{Liong_2015_CVPR} at all code lengths, in both mAP and precision of Hamming radius.


\vspace{-0.2cm}
\section{Supervised Hashing with Binary Deep Neural Network (SH-BDNN)}
\label{sec:SH-BDNN}
\vspace{-0.2cm}
In order to enhance the discriminative power of the binary codes, we extend UH-BDNN to supervised hashing by leveraging the label information. There are several approaches proposed to leverage the label information, leading to different criteria on binary codes. In~\cite{DBLP:journals/pami/StrechaBBF12,minhdo_hash2014}, binary codes are learned such that they minimize Hamming distance between samples belonging to same class, while maximizing the Hamming distance between samples belonging to different classes. In~\cite{Shen_2015_CVPR}, the binary codes are learned such that they are optimal for linear classification.

In this work, in order to leverage the label information, we follow the approach proposed in Kernel-based Supervised Hashing (KSH)~\cite{CVPR12:Hashing}. The benefit of this approach is  
that it directly encourages the Hamming distances between binary codes of within-class samples equal to $0$, and the Hamming distances between binary codes of between-class samples equal to $L$. In the other words, it tries to perfectly preserve the semantic similarity. To achieve this goal, it enforces that the 
Hamming distance between learned binary codes has to highly correlate with the pre-computed pairwise label matrix.


In general, the network structure of SH-BDNN is similar to UH-BDNN, excepting that the last layer preserving reconstruction of UH-BDNN is removed. The layer $n-1$ in UH-BDNN becomes the last layer in SH-BDNN. All desirable properties, i.e. semantic similarity preserving, independence, and balance, in SH-BDNN are constrained on the outputs of its last layer. 

\vspace{-0.3cm}
\subsection{Formulation of SH-BDNN}
We define the pairwise label matrix $\S$ as
\small
\begin{equation}
\S_{ij} = \left\{ \begin{array}{ll}
1 & \textrm{if $\x_i$ and $\x_j$ are same class}\\
-1 & \textrm{if $\x_i$ and $\x_j$ are not same class}
\end{array} \right.
\label{eq:S}
\end{equation}
\normalsize
To achieve the semantic similarity preserving property, we learn the binary codes such that the Hamming distance between learned binary codes highly correlates with the matrix $\S$, i.e.,  we want to minimize the quantity  $\norm{\frac{1}{L} (\H^{(n)})^T\H^{(n)} - \S}^2$.
In addition, to achieve the independence and balance properties of codes, we want to minimize the quantities $\norm{\frac{1}{m}\H^{(n)}(\H^{(n)})^T-\I}^2$ and $\norm{\H^{(n)}\1_{m\times 1}}^2$.

Follow the same reformulation and relaxation as UH-BDNN (Sec.~\ref{subsec:formular_un}), we solve the following constrained optimization which ensures the binary constraint, the semantic similarity preserving, the independence, and the balance properties of codes

\vspace{-0.4cm}\footnotesize
\begin{eqnarray}
\min_{\W,\cc,\B} J &=& \frac{1}{2m}\norm{\frac{1}{L} (\H^{(n)})^T\H^{(n)} - \S}^2 +\frac{\lambda_1}{2}\sum_{l=1}^{n-1} \norm{\W^{(l)}}^2 + \frac{\lambda_2}{2m}\norm{\H^{(n)}-\B}^2 \nonumber \\
{}&&+\frac{\lambda_3}{2}\norm{\frac{1}{m}\H^{(n)}(\H^{(n)})^T-\I}^2 +\frac{\lambda_4}{2m}\norm{\H^{(n)}\1_{m\times 1}}^2
 \label{eq:obj_sup2}
\end{eqnarray}
\begin{equation}
\hspace{-1.5cm}\textrm{s.t. }\B \in \{-1,1\}^{L\times m} \label{eq:binary_H1_sup2}
\end{equation}
\normalsize
(\ref{eq:obj_sup2}) under constraint (\ref{eq:binary_H1_sup2}) is our formulation for supervised hashing. The main difference in formulation between UH-BDNN~(\ref{eq:obj_5}) and  SH-BDNN~(\ref{eq:obj_sup2}) is that the reconstruction term preserving the neighbor similarity in UH-BDNN~(\ref{eq:obj_5}) is replaced by the term preserving the label similarity in SH-BDNN~(\ref{eq:obj_sup2}).

\vspace{-0.3cm}
\subsection{Optimization}
\begin{algorithm}[!t]
	\scriptsize
	\caption{Supervised Hashing with Binary Deep Neural Network (SH-BDNN)}
	\begin{algorithmic}[1] 
		\Require 
			\Statex $\X = \{\x_i\}_{i=1}^{m} \in \R^{D\times m}$: training data; $\Y \in R^{m\times 1}$: training label vector; $L$: code length; $T$: maximum iteration number; $n$: number of layers; $\{s_l\}_{l=2}^{n}$: number of units of layers $2 \to n$ (note: $s_n = L$); $\lambda_1, \lambda_2, \lambda_3, \lambda_4$.
		\Ensure 
			\Statex 
			Parameters $\{\W^{(l)},\cc^{(l)}\}_{l=1}^{n-1}$
			\Statex 
			\State Compute pairwise label matrix $\S$ using~(\ref{eq:S}).
			\State Initialize $\B_{(0)} \in \{-1,1\}^{L\times m}$ using ITQ~\cite{DBLP:conf/cvpr/GongL11}
			\State Initialize $\{\cc^{(l)}\}_{l=1}^{n-1} = \mathbf{0}_{s_{l+1}\times 1}$. Initialize $\{\W^{(l)}\}_{l=1}^{n-1}$ by getting the top $s_{l+1}$ eigenvectors from the covariance matrix of $\H^{(l)}$. 
			\State Fix $\B_{(0)}$, compute $(\W,\cc)_{(0)}$ with $(\W,\cc)$ step using initialized $\{\W^{(l)},\cc^{(l)}\}_{l=1}^{n-1}$ (line 3) as starting point for L-BFGS.
			\For{$t = 1 \to T$}
				\State Fix $(\W,\cc)_{(t-1)}$, compute $\B_{(t)}$ with $\B$ step
				\State Fix $\B_{(t)}$, compute $(\W,\cc)_{(t)}$ with $(\W, \cc)$ step using $(\W,\cc)_{(t-1)}$ as starting point for L-BFGS.
			\EndFor
			\State Return 
			$(\W,\cc)_{(T)}$
    \end{algorithmic}
    \label{alg2}
\end{algorithm}
In order to solve~(\ref{eq:obj_sup2}) under constraint~(\ref{eq:binary_H1_sup2}), we alternating optimize over $(\W,\cc)$ and $\B$.

\vspace{-0.3cm}
\subsubsection{$(\W,\cc)$ step}
\label{subsub:W_step_sup}
When fixing $\B$, (\ref{eq:obj_sup2}) becomes unconstrained optimization. We used \textit{L-BFGS}~\cite{Liu89onthe} optimizer with backpropagation for solving. The gradient of objective function $J$ (\ref{eq:obj_sup2}) w.r.t. different parameters are computed as follows.

Let us define

\vspace{-0.4cm}\footnotesize
\begin{eqnarray}
\Delta^{(n)} &=& \left[ \frac{1}{mL}\H^{(n)}\left( \V+\V^T \right)+\frac{\lambda_2}{m}\left( \H^{(n)}-\B \right) +\frac{2\lambda_3}{m}\left( \frac{1}{m}\H^{(n)}(\H^{(n)})^T - \I\right)\H^{(n)} \right. \nonumber \\
 {}&& \left. +\frac{\lambda_4}{m}\left( \H^{(n)}\1_{m\times m} \right) \right]\odot f^{(n)'}(\Z^{(n)})
\end{eqnarray}
\normalsize
where \small$\V = \frac{1}{L}(\H^{(n)})^T\H^{(n)} - \S$.\normalsize

\vspace{-0.2cm}\footnotesize
\begin{equation}
\Delta^{(l)} = \left( (\W^{(l)})^T\Delta^{(l+1)} \right) \odot f^{(l)'}(\Z^{(l)}),\forall l = n-1,\cdots,2
\end{equation}
\normalsize
where $\odot$ denotes Hadamard product; \small$\Z^{(l)} = \W^{(l-1)}\H^{(l-1)} + \cc^{(l-1)} \1_{1\times m}$, $l=2,\cdots,n$.\normalsize

Then $\forall l = n-1,\cdots,1$, we have

\vspace{-0.2cm}\footnotesize
\begin{equation}
\frac{\partial J}{\partial \W^{(l)}} = \Delta^{(l+1)}(\H^{(l)})^T +\lambda_1\W^{(l)}
\end{equation}
\begin{equation}
\frac{\partial J}{\partial \cc^{(l)}} = \Delta^{(l+1)}\1_{m\times 1}
\end{equation}
\normalsize
\vspace{-0.3cm}
\subsubsection{$\B$ step}
\label{subsub:B_step_sup}
When fixing $(\W,\cc)$, we can rewrite problem~(\ref{eq:obj_sup2}) as 

\vspace{-0.3cm}\footnotesize
\begin{equation}
\min_{\B} J =  \norm{\H^{(n)}-\B}^2
\label{eq:obj_sup3}
\end{equation}
\vspace{-0.2cm}
\begin{equation}
\textrm{s.t.\ } \B \in \{-1,1\}^{L\times m} \label{eq:binary_H3}
\end{equation}
\normalsize
It is easy to see that the optimal solution for~(\ref{eq:obj_sup3}) under constraint~(\ref{eq:binary_H3}) is $\B = sgn(\H^{(n)})$.

The proposed SH-BDNN method is summarized in Algorithm~\ref{alg2}. In the Algorithm~\ref{alg2}, $\B_{(t)}$ and $(\W,\cc)_{(t)}$ are values of $\B$ and $\{\W^{(l)},\cc^{(l)}\}_{l=1}^{n-1}$ at iteration $t$.
\vspace{-0.4cm}
\section{Evaluation of Supervised Hashing with Binary Deep Neural Network (SH-BDNN)}
\label{sec:eva_SH-BDNN}
\vspace{-0.2cm}
This section evaluates the proposed SH-BDNN and compares it with following state-of-the-art supervised hashing methods: Supervised Discrete Hashing (SDH)~\cite{Shen_2015_CVPR}, ITQ-CCA~\cite{DBLP:conf/cvpr/GongL11}, Kernel-based Supervised Hashing (KSH)~\cite{CVPR12:Hashing}, Binary Reconstructive Embedding (BRE)~\cite{Kulis_learningto}. For all compared methods, we use the implementation and the suggested parameters provided by the authors.
\vspace{-0.3cm}
\subsection{Dataset, evaluation protocol, and implementation note}
\subsubsection{Dataset} 
We evaluate and compare methods on CIFAR-10 and MNIST datasets. The descriptions of these datasets are presented in section~\ref{subsec:data-imp-eva}.
\vspace{-0.4cm} 
\subsubsection{Evaluation protocol} 
Follow the literature~\cite{Shen_2015_CVPR,DBLP:conf/cvpr/GongL11,CVPR12:Hashing}, we report the retrieval results in two metrics: 1) mean Average Precision (mAP) and 2) precision of Hamming radius $2$ (precision$@2$).

\vspace{-0.4cm}
\subsubsection{Implementation note} The network configuration is same as UH-BDNN excepting the final layer is removed. The values of parameters $\lambda_1$, $\lambda_2$, $\lambda_3$ and $\lambda_4$ are empirically set using cross validation as $10^{-3}$, $5$, $1$ and $10^{-4}$, respectively. The max iteration number $T$ is empirically set to 5. 

Follow the settings in ITQ-CCA~\cite{DBLP:conf/cvpr/GongL11}, SDH~\cite{Shen_2015_CVPR},  all training samples are used in the learning for these two methods. For SH-BDNN, KSH~\cite{CVPR12:Hashing} and BRE~\cite{Kulis_learningto} where label information is leveraged by the pairwise label matrix, we randomly select $3,000$ training samples from each class and use 
them for learning. The ground truths of queries are defined by the class labels from the datasets.

\vspace{-0.3cm}
\subsection{Retrieval results}
\begin{figure*}[!t]
\vspace{-1em}
\centering
\subfigure[CIFAR10]{
       \includegraphics[scale=0.30]{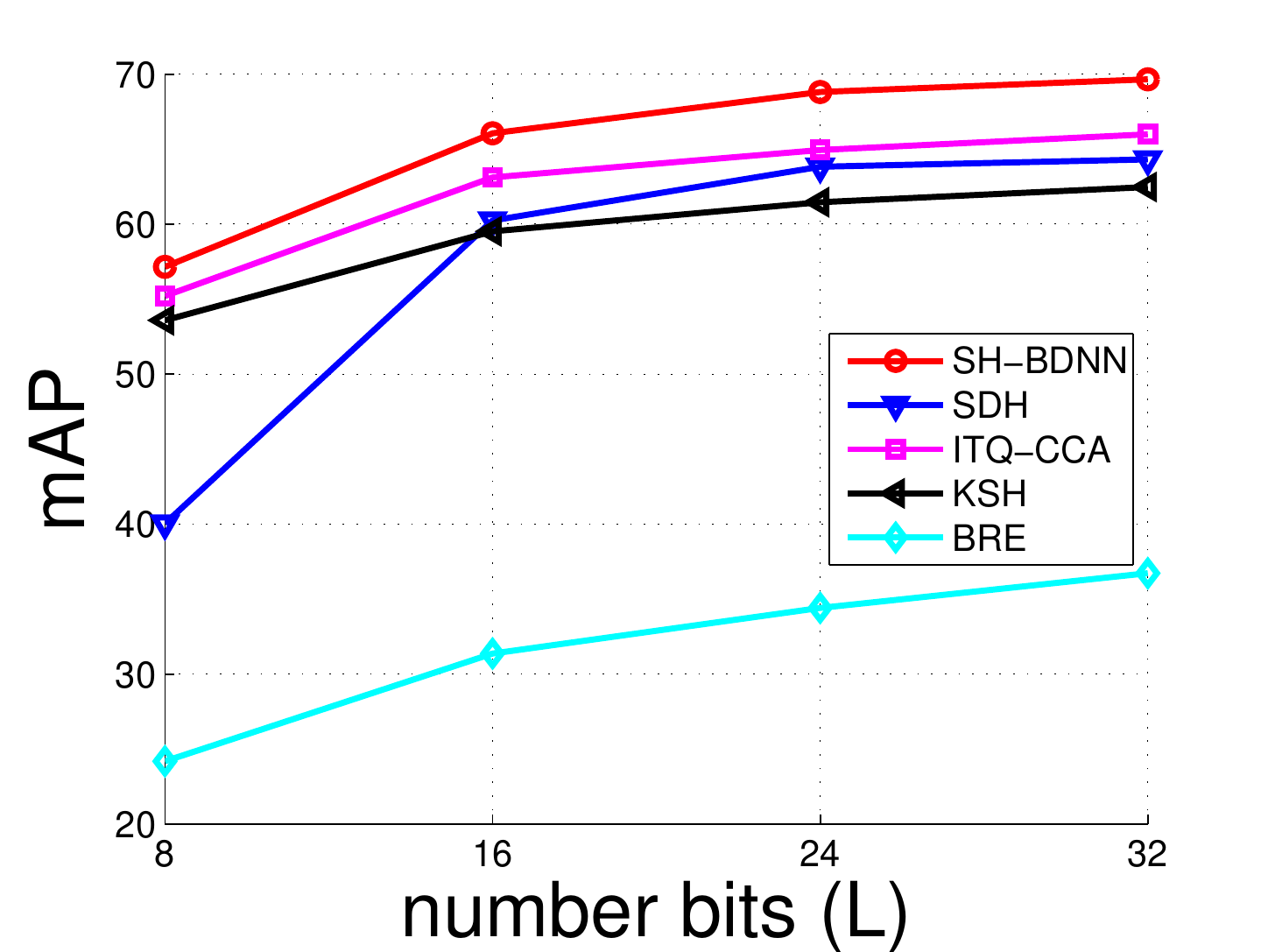}
       \label{fig:cifar_mAP_sup}
}
\subfigure[MNIST]{
       \includegraphics[scale=0.30]{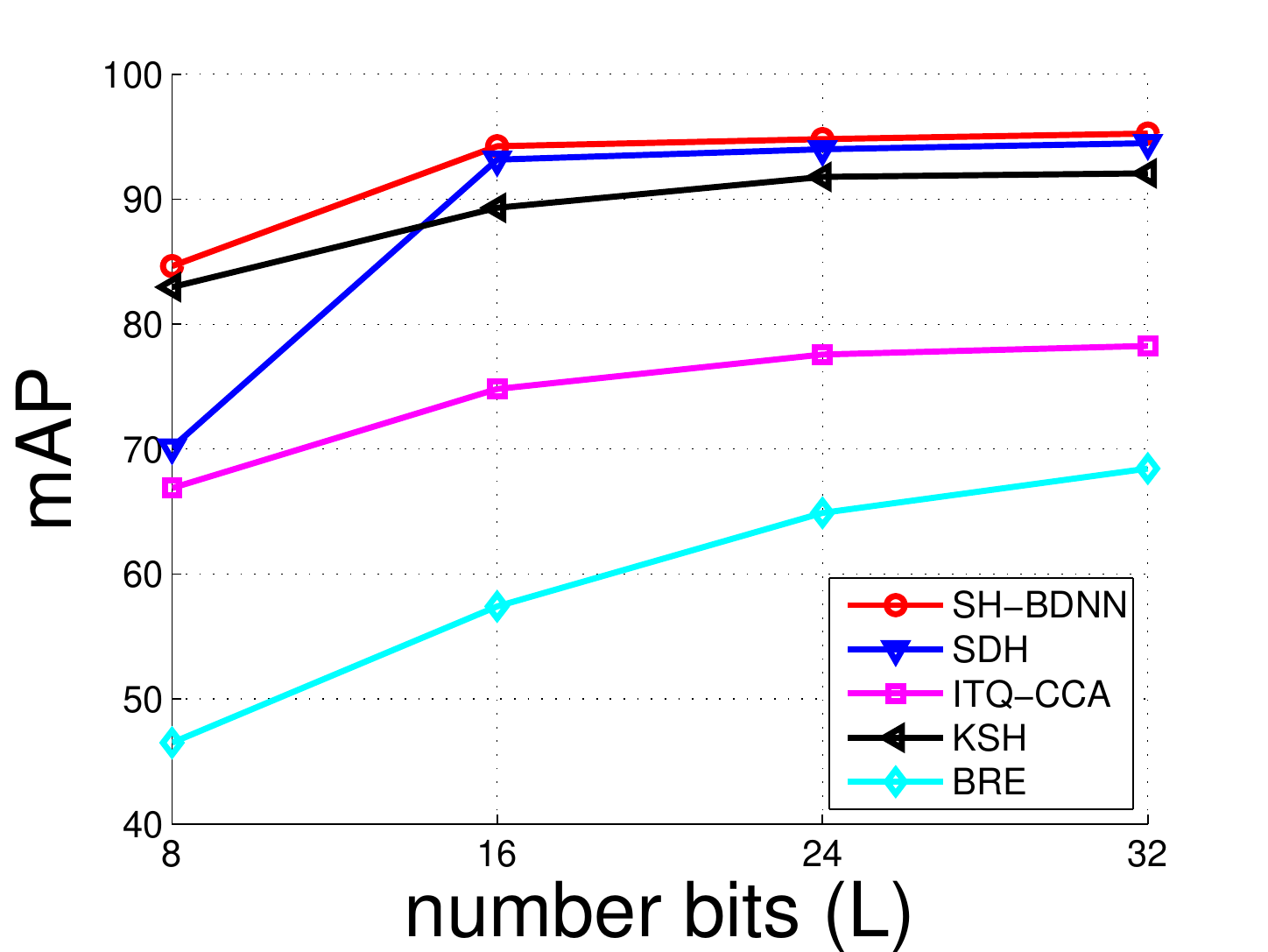} 
       \label{fig:mnist_mAP_sup}
}
\vspace{-0.2cm}
\caption[]{mAP comparison between SH-BDNN and state-of-the-art supervised hashing methods on CIFAR10 and MNIST.}
\label{fig:cifar10_mnist_sup}
\end{figure*}
\begin{table}[!t]
   \centering
   \footnotesize
   \caption{Precision at Hamming distance $r=2$ comparison between SH-BDNN and state-of-the-art supervised hashing methods on CIFAR10 and MNIST.}
    \begin{tabular}{|l|c| c| c| c|c| c| c| c|}
		\hline
	  \multirow{2}{*}{} & \multicolumn{4}{|c|}{CIFAR10} & \multicolumn{4}{|c|}{MNIST} \\
\cline{1-9}	$L$    		          &8 &16 &24 &32    		    &8 &16 &24 &32     \\ \hline 
SH-BDNN						      &54.12 &67.32 &69.36 &69.62   &84.26 &94.67 &94.69 &95.51 
		\\ \hline
SDH\cite{Shen_2015_CVPR} 		  &31.60 &62.23 &67.65 &67.63   &36.49  &93.00 &93.98 &94.43 
		\\ \hline  
ITQ-CCA\cite{DBLP:conf/cvpr/GongL11}&49.14 &65.68 &67.47 &67.19  &54.35 &79.99 &84.12 &84.57
		\\ \hline		                                  
KSH\cite{CVPR12:Hashing}		  &44.81 &64.08 &67.01 &65.76   &68.07  &90.79 &92.86 &92.41  
		\\ \hline
BRE\cite{Kulis_learningto}        &23.84 &41.11 &47.98 &44.89   &37.67  &69.80 &83.24 &84.61
		\\ \hline
	  \end{tabular}
	  \label{tab:soa-sup-cifar10-mnist-pre}
	  \vspace{-0.2cm}
\end{table}
On CIFAR10 dataset, Fig.~\ref{fig:cifar_mAP_sup} and Table~\ref{tab:soa-sup-cifar10-mnist-pre} clearly show the proposed SH-BDNN outperforms all compared methods by a fair margin at all code lengths in both mAP and precision@2. 


On MNIST dataset, Fig.~\ref{fig:mnist_mAP_sup} and Table~\ref{tab:soa-sup-cifar10-mnist-pre} show the proposed SH-BDNN significantly outperforms the current state-of-the-art SDH at low code length, i.e., $L=8$. When $L$ increases, SH-BDNN and SDH~\cite{Shen_2015_CVPR} achieve similar performance. In comparison to remaining methods, i.e., KSH~\cite{CVPR12:Hashing}, ITQ-CCA~\cite{DBLP:conf/cvpr/GongL11}, BRE~\cite{Kulis_learningto}, SH-BDNN outperforms these methods by a large margin in both mAP and precision@2.\\
\textit{Comparison with CNN-based hashing methods~\cite{DBLP:conf/cvpr/ZhaoHWT15,DBLP:journals/tip/ZhangLZZZ15}} 
We also compare our proposed SH-BDNN to the very recent CNN-based supervised hashing methods:  Deep Semantic Ranking Hashing (DSRH)~\cite{DBLP:conf/cvpr/ZhaoHWT15} and Deep Regularized Similarity Comparison Hashing (DRSCH)~\cite{DBLP:journals/tip/ZhangLZZZ15}. 
Note that the focus of~\cite{DBLP:conf/cvpr/ZhaoHWT15,DBLP:journals/tip/ZhangLZZZ15} are different from ours: 
in~\cite{DBLP:conf/cvpr/ZhaoHWT15,DBLP:journals/tip/ZhangLZZZ15}, the authors focus on a  framework in which the image features and hash codes are 
{\em jointly}
learned by combining CNN layers (image feature extraction) and binary mapping layer into a single model.
On the other hand, our work focuses on only the binary mapping layer given some image feature.
Note that in~\cite{DBLP:conf/cvpr/ZhaoHWT15,DBLP:journals/tip/ZhangLZZZ15}, 
their binary mapping layer  only applies a simple operation, i.e., an approximation of $sgn$ function (i.e., $logistic$~\cite{DBLP:conf/cvpr/ZhaoHWT15}, $tanh$~\cite{DBLP:journals/tip/ZhangLZZZ15}), on CNN features for achieving the approximated binary codes. 
Our SH-BDNN advances~\cite{DBLP:conf/cvpr/ZhaoHWT15,DBLP:journals/tip/ZhangLZZZ15} in the way to map the image features to the binary codes (which is our main focus). Given the image features (i.e., pre-trained CNN features), 
we apply multiple transformations on these features; we constrain one layer to directly output the binary code, without involving $sgn$ function. Furthermore, our learned codes ensure good properties, i.e. independence and balance, while DRSCH~\cite{DBLP:journals/tip/ZhangLZZZ15} does not consider such properties, and DSRH~\cite{DBLP:conf/cvpr/ZhaoHWT15} only considers the balance of codes. 

We follow strictly the comparison setting in~\cite{DBLP:conf/cvpr/ZhaoHWT15,DBLP:journals/tip/ZhangLZZZ15}.
In~\cite{DBLP:conf/cvpr/ZhaoHWT15,DBLP:journals/tip/ZhangLZZZ15}, when comparing their CNN-based hashing to other non CNN-based hashing methods, the authors use pre-trained CNN features (e.g. AlexNet~\cite{jia2014caffe}, DeCAF~\cite{DBLP:conf/icml/DonahueJVHZTD14}) as input for other methods. Follow that setting, we use AlexNet features \cite{jia2014caffe} as input for SH-BDNN  
when comparing SH-BDNN to~\cite{DBLP:conf/cvpr/ZhaoHWT15,DBLP:journals/tip/ZhangLZZZ15}. We set up the experiments on CIFAR10 similar to~\cite{DBLP:journals/tip/ZhangLZZZ15}, i.e., the query set contains 10K images (1K images per class) randomly sampled from the dataset; the rest 50K image are used as the training set; 
in the testing step, each query image is searched within the query set itself by applying the leave-one-out procedure.

\begin{table}[!t]
   \centering
   \footnotesize
   \vspace{-0.2em}
   \caption{Comparison between SH-BDNN and CNN-based hashing DSRH~\cite{DBLP:conf/cvpr/ZhaoHWT15}, DRSCH~\cite{DBLP:journals/tip/ZhangLZZZ15} on CIFAR10. The results of DSRH and DRSCH are cited from~\cite{DBLP:journals/tip/ZhangLZZZ15}.}
   \vspace{-0.2em}
    \begin{tabular}{|c|c|c|c|c|c|c|c|c|}
		\hline
	  \multirow{2}{*}{} & \multicolumn{4}{|c|}{mAP} & \multicolumn{4}{|c|}{precison@2} \\
\cline{1-9}	$L$    		          &16 &24 &32 &48    		  &16 &24 &32 &48     \\ \hline 
SH-BDNN				&64.30 &65.21 &66.22 &66.53   &56.87 &58.67 &58.80 &58.42 \\ \hline
DRSCH\cite{DBLP:journals/tip/ZhangLZZZ15}&61.46 &62.19 &62.87 &63.05 &52.34 &53.07 &52.31 &52.03\\ \hline  
DSRH\cite{DBLP:conf/cvpr/ZhaoHWT15}&60.84 &61.08 &61.74 &61.77 &50.36 &52.45 &50.37 &49.38\\ \hline
	  \end{tabular}
	  \label{tab:soa-sup-cvpr15-tip}
	  \vspace{-0.4cm}
\end{table}
The comparative results between the proposed SH-BDNN and DSRH~\cite{DBLP:conf/cvpr/ZhaoHWT15}, DRSCH~\cite{DBLP:journals/tip/ZhangLZZZ15}, presented in Table~\ref{tab:soa-sup-cvpr15-tip}, clearly show that at the same code length, the proposed SH-BDNN outperforms~\cite{DBLP:conf/cvpr/ZhaoHWT15} and~\cite{DBLP:journals/tip/ZhangLZZZ15} in both mAP and precision@2.

\vspace{-0.3cm}
\section{Conclusion}
\label{sec:conclusion}
\vspace{-0.2cm}
In this paper, we propose UH-BDNN and SH-BDNN for unsupervised and supervised hashing.
Our network designs constrain to directly produce binary codes at one layer.  Our models ensure good properties for codes:  similarity preserving, independence and balance. Solid experimental results on three benchmark datasets show that the proposed methods 
compare favorably with the state of the art. 


\bibliographystyle{splncs}
\bibliography{hash}
\end{document}